\def\year{2021}\relax
\documentclass[letterpaper]{article}
\usepackage{aaai21} 
\usepackage{times} 
\usepackage{helvet} 
\usepackage{courier} 
\usepackage[hyphens]{url} 
\usepackage{graphicx} 
\usepackage{graphicx}  
\usepackage{natbib}
\usepackage{varwidth}
\usepackage{caption}

\usepackage{bm}
\usepackage{algorithm}
\usepackage{multirow}
\usepackage{multicol}
\usepackage[noend]{algpseudocode}
\usepackage{amsthm,amssymb, amsmath}
\usepackage[utf8]{inputenc} 
\usepackage[T1]{fontenc}    
\usepackage{hyperref}       
\usepackage{url}            
\usepackage{booktabs}       
\usepackage{amsfonts}       
\usepackage{nicefrac}       
\usepackage{microtype}      
\usepackage{color}          
\usepackage{enumitem}
\usepackage{bbm}
\usepackage{graphicx}

\DeclareMathOperator*{\argmax}{arg\,max}

\title{You Only Query Once: Effective Black Box Adversarial Attacks with Minimal Repeated Queries}
\author{
	Devin Willmott\textsuperscript{\rm 1},
	Anit Kumar Sahu\textsuperscript{\rm 2},
	Fatemeh Sheikholeslami\textsuperscript{\rm 1},
	Filipe Condessa\textsuperscript{\rm 1}\&
	Zico Kolter\textsuperscript{\rm 1,3}
}

\affiliations{
	\textsuperscript{\rm 1}Bosch Center for AI,
	\textsuperscript{\rm 2}Amazon Alexa AI (work completed while at Bosch Center for AI),
	\textsuperscript{\rm 3}Carnegie Mellon University
}

\begin{document}
	\maketitle
	\begin{abstract}
		Researchers have repeatedly shown that it is possible to craft adversarial attacks on deep classifiers (small perturbations that significantly change the class label), even in the ``black-box'' setting where one only has query access to the classifier.  However, all prior work in the black-box setting attacks the classifier by repeatedly querying the \emph{same} image with minor modifications, usually thousands of times or more, making it easy for defenders to detect an ensuing attack.  In this work, we instead show that it is possible to craft (universal) adversarial perturbations in the black-box setting by querying a sequence of \emph{different} images only once. This attack prevents detection from high number of similar queries and produces a perturbation that causes misclassification when applied to \emph{any} input to the classifier. In experiments, we show that attacks that adhere to this restriction can produce untargeted adversarial perturbations that fool the vast majority of MNIST and CIFAR-10 classifier inputs, as well as in excess of $60-70\%$ of inputs on ImageNet classifiers. In the targeted setting, we exhibit targeted black-box universal attacks on ImageNet classifiers with success rates above $20\%$ when only allowed one query per image, and $66\%$ when allowed two queries per image.
	\end{abstract}
	
	\section{Introduction}
	
	It has repeatedly been shown that deep networks are sensitive to adversarial attacks \cite{goodfellow2014explaining,autozoom,carlini2017,zoo}, small changes in the input to classifier that can drastically change the output.  And although the majority of adversarial attacks operate in the ``white-box'' setting (i.e., assuming full knowledge of the classifier, including the ability to backpropagate through it) it is also well-established that these attacks can transfer to the ``black-box'' setting, where one only has query access to the classifier.  However, existing black-box attacks suffer from two distinct limitations.  First, they require repeatedly querying the classifier (often many hundreds or thousands of times) on the \emph{same} image or very similar images, with minor variations introduced to assess the effect of local perturbations; this makes such attacks relatively easy to detect in an operational setting (by simply flagging queries that repeatedly request classifications of images that are very close in some norm).  Second, partially because of this setup, all existing black box attacks we are aware of generate an adversarial example specific to a given image, which does not transfer to other images.
	
	In this paper, we propose and demonstrate an approach that overcomes these issues.  We show for the first time that it is possible to create black-box attacks by querying many \emph{different} images a small number of times each (including the limiting case of querying each image just once).  By its nature, this also creates a perturbation that is \emph{universal} \cite{moosavi2017universal}, in the sense that the exact same perturbation can be applied to any input image and still consistently lead to misclassification.  Methodologically, the approaches we use are straightforward: we employ a two-sided empirical gradient estimate in the two-queries-per-image setting, and use the covariance matrix adaptation~(CMA-ES) algorithm~\cite{hansen2016cma} for the one-query-per-image setting.  From a practical standpoint, however, the work demonstrates that this very general form of attack is possible even with very limited interaction with a black box model.
	
	
	We apply our methods to learn universal adversarial perturbations in various experimental settings, including attacks on MNIST, CIFAR-10, and ImageNet classifiers. We show that these attacks can, in an entirely black-box fashion, produce untargeted adversarial perturbations that fool the vast majority of MNIST and CIFAR-10 classifier inputs, and achieve attack success rates above $60-70\%$ on many ImageNet classifiers. With appropriate adjustments to attack parameters and the use of additional unlabeled data for additional network queries, our attacks are also suitable for the targeted setting: we exhibit targeted black-box universal attacks on the ResNet-50 ImageNet classifier with success rates above $20\%$ when only allowed one query per image, and above $66\%$ when allowed two queries per image.
	
	\section{Related Work}
	
	Earliest work exhibiting the efficacy of adversarial attacks on deep classifiers focused on generating a perturbation that causes a network to misclassify a particular input using gradient-based methods~\cite{szegedy2013intriguing,goodfellow2014explaining,biggio2013evasion}. Since these early papers, the field has grown tremendously to consider numerous variations in the goals and capabilities of the attacker.  Two such variations are \textit{universal attacks} \cite{moosavi2017universal}, which seek to produce a single perturbation $\boldsymbol{\delta}$ that is likely to cause misclassification when applied to \textit{any} network input instead of a particular image $\boldsymbol{x}$, and \textit{black-box attacks}~\cite{zoo}, which assume no knowledge of the architecture or parameters of the network.  This paper indeed focuses on the intersection of these two settings: generating universal perturbations for an attacker which only has black-box query access, which has not been considered previously.

	\textbf{Universal Adversarial Perturbations.} The first universal adversarial attack on deep learning classifiers was presented in \cite{moosavi2017universal}, where a universal perturbation $\boldsymbol{\delta}$ is generated by iterating over a set of input images one or more times and using the DeepFool attack \cite{moosavi2016deepfool} on $\boldsymbol{x}$, which finds the minimum-norm perturbation that sends $\boldsymbol{x}$ to the decision boundary. Subsequent universal adversarial attacks have followed similar approaches to this, either replacing DeepFool with another single input attack \cite{hirano2019simple} or, in other cases~\cite{hayes2018learning,reddy2018nag}, replacing direct updates to a single perturbation with training steps for a generative model from which perturbations may be sampled.
	
	It is worth noting that all of the above papers rely on access to the gradients of the model being attacked. Among these, many \cite{moosavi2017universal,hayes2018learning,reddy2018nag} also consider how often adversarial perturbations generated on one network cause misclassification on other networks trained on the same task. These approaches are part of a category of \textit{transfer-based attacks}, where perturbations are first found by attacking a surrogate network and then applied to a second target network. These attacks come with their own benefits and downsides; most notably, they require first training the surrogate network, which may be expensive and require large amounts of labeled data.
	
	\textbf{Black Box Attacks.} Following papers that popularized the notion that gradient-based methods are effective at fooling deep classifiers, a more recent and burgeoning area of research has revealed that such vulnerabilities remain even in the black-box setting \cite{autozoom,zoo,bandits,nes}, where one only has query access to the classifier. Nearly all of these methods draw directly from a wide array of existing derivative-free optimization (DFO) techniques, including combinatorial optimization \cite{parsimonious}, Bayesian optimization \cite{bayesopt,bayesopt_tutorial} and more. Though we consider the modified setting of universal attacks, the DFO methods that we use in this work have also been explored in other black-box attack algorithms; namely, the evolution strategy we use is related to approaches found in \cite{nes}, which uses the natural evolution strategy (NES) algorithm \cite{wierstra2014natural}, and \cite{meunier2019yet}, which uses a similar version of CMA-ES used here. Finite difference based methods \cite{zoo,nes,bandits} have also been used extensively to generate black-box adversarial attacks.
	
	
	Beyond optimization techniques, our attack also incorporates prior works' insight into the nature of adversarial attacks. In particular, some \cite{bandits,sharma2019effectiveness} have highlighted the efficacy of low-dimensional adversarial perturbations for black-box attacks, which inspires our tiling dimension reduction technique.
	
	\textbf{Targeted vs. Untargeted Attacks.} Another important distinction in adversarial attacks is whether the attacker aims at reducing the classification accuracy by merely changing the output to a random class, a.k.a. untargeted attacks, or if the goal is to shift the classification outcome onto a particular \emph{target} class \cite{carlini2017}. The fragility of neural networks have lead the former to be a fairly simple task, as even single step attacks can achieve a high success rate in white-box settings. On the other hand, targeted attacks, where attacks try to cause the network to predict a specified target class regardless of input, are considerably more difficult, particularly in classification tasks with large number of classes. This leads to a much higher number of per image queries in black box settings \cite{autozoom}.

	\subsection{Preliminaries on Adversarial Attacks}
	
	Adversarial examples are modeled as inputs with carefully crafted additive perturbations that cause the classifier to misclassify  the input. Typically this is done by perturbing the input while still ensuring minimal visual distortion, often achieved by imposing constraints on the $\ell_p$-norm of the perturbation, e.g., $\|\boldsymbol{\delta}\|_p \leq \epsilon_p$ for $p=1,2, \infty$. Throughout this paper, we specifically focus on the case where the constraint imposed is in $\ell_{\infty}$ norm.

	Formally, we define a classifier $C:\mathcal{X}\to\mathbb{R}^K$ where $\mathbf{x}\in\mathcal{X}$ is the input image, $y\in\mathcal{Y}=\{1,2,..., K\}$ is the output space, the outputs of the classifier are the logit values for each class label respectively.  The quality of a classifier is assessed via a classification loss $L(C(\mathbf{x}),y)$, which we typically take to be the cross entropy loss
	\begin{equation}
	\label{eq:cross-entropy}
	L(C(\mathbf{x}),y) = \log\sum_{k=1}^K \exp(C(\mathbf{x})_k) - C(\mathbf{x})_y
	\end{equation}
	and which we will abbreviate as $L(\mathbf{x},y)$.  The attacker's objective can then be formalized as finding a perturbation $\bm{\delta}$ that maximizes this loss for a particular input $\mathbf{x}$, subject to the constraint ${  \|\bm{\delta}\|_p \leq \epsilon_p}$. Thus, we define the adversarial perturbation to be
	\begin{align}
	\label{eq:opt_single}
	\bm{\delta}^*(\mathbf{x}) := \argmax_{\bm{\delta} :  \|\bm{\delta}\|_p \leq \epsilon_p} L(\mathbf{x}+\bm{\delta},y).
	\end{align}
	
	
	This formulation however, aims at finding  a {\emph {separate}} adversarial perturbation for every input $\mathbf{x}$, assuming the feasibility of using different perturbations for each input. On the other hand, most physically realizable attacks, implemented via \emph{stickers} or \emph{patches} \citep{brown2017adversarsial,li2019adversarial}, are restricted to having the same perturbation across all input images. This motivates the search for a single \emph{universal} perturbation that can disrupt classification over the entire distribution of the input $\mathbf{x}\in\mathcal{X}$. 
	
	Such a perturbation can be obtained by solving the following reformulation of the objective:
	\begin{align}
	\label{eq:opt1}
	\bm{\delta}^* = \argmax_{\bm{\delta} :  \|\bm{\delta}\|_p \leq \epsilon_p} \mathbb{E}_{\{\mathbf{x},y\}\sim\mathcal{P}_{\mathcal{X}}}[L(\mathbf{x}+\bm{\delta},y)],
	\end{align}
	where the expectation is taken over image-label pairs $\{\mathbf{x},y\}$ drawn from the classifier's underlying data distribution $\mathcal{P}_{\mathcal{X}}$. Since $\mathcal{P}_{\mathcal{X}}$ is typically unavailable, one in practice resorts to its empirical mean approximation, rendering the objective to be as 
	\begin{align}
	\label{eq:opt2}
	\hat{\bm{\delta}} = \argmax_{\bm{\delta} :  \|\bm{\delta}\|_p \leq \epsilon_p} \dfrac{1}{N} \sum_{n=1}^N L(\mathbf{x}_n+\bm{\delta},y_n).
	\end{align}
	where $\{\mathbf{x}_n, y_n\}_{n=1}^N$ is a set of image-label pairs in the joint distribution.
	
	Given a white-box oracle that has full access to the classifier as well as the gradient of the objective through back-propagation, such {\emph {universal}} perturbations can be obtained by the well-known iterative (class of) projected gradient ascent updates \cite{madry2018towards}, with the update at iteration $t$ given by
	\begin{align}
	\label{eq:pgd_update}
	&\boldsymbol{\delta}^{(t+1)} \leftarrow  \mathcal{P}_{B(0,\epsilon)}^\infty (\boldsymbol{\delta}^{(t)} + \eta \; \text{sign}(\mathbf{\Bar{g}}^{(t)} )),\nonumber\\&\text{where}~~\Bar{\mathbf{g}}^{(t)} := \dfrac{1}{B}  \sum_{b=1}^{B}\nabla_{\bm{\delta}} L(\mathbf{x}_b+\boldsymbol{\delta}^{(t)},y_b).
	\end{align}
	where $\eta$ is the stepsize, $\mathcal{P}_{B(0,\epsilon)}^\infty$ denotes projection onto the $\ell_\infty$ ball, 
	and $\bar{\mathbf{g}}^{(t)}$ is the gradient of the loss function with respect to the perturbation.
	While setting $B=N$ corresponds to the full gradient of the objective in \eqref{eq:opt2},
	a variety of techniques, such as stochastic alternatives with $B<N$ or incorporation of momentum, may be used to speed up this optimization. 
	
	In realistic settings, the classifier is often unknown to the attackers, making such approaches that rely on full knowledge of the gradient inapplicable. Thus, in order to obtain realizable attacks one needs to consider {\em black-box} settings, in which the architecture of the classifier  as well as its parameters are hidden from the attacker \cite{autozoom,zoo,bandits,nes}. Typically, in such black-box settings, the adversary is modeled as only having access to a zeroth order oracle that supplies the loss function value $L(\mathbf{x},y)$.
	In spite of the information constraints and typically high dimensional inputs, it has been shown that black-box attacks can be fairly successful in addressing such cases, but this success comes at the price of high numbers of per-image queries \citep{bandits,nes,parsimonious}. However, finding effective targeted and untargeted universal adversarial perturbations in a black-box setting under stringent constraints on the number of per-image queries is still an uncharted area of research, and is the subject of this work.

	\section{Universal Perturbations with Limited Queries}
	The necessity of querying many minute variations of a single input image to successfully produce an adversarial perturbation for that image significantly is a significant hindrance to adversarial attacks in a real-world scenario. A defender with knowledge of this limitation may use simple criteria to detect attacks and then take appropriate action -- for example, they may lock out users who query slight variations of the same image more than a certain number of times. It is thus natural to ask whether it is possible for a black-box attack to evade such detection. To determine this, we consider how to best construct adversarial perturbations while keeping the number of queries at or near any particular image low. More precisely, given an $\ell_{\infty}$ constraint $\epsilon$ to which the perturbation must adhere, we allow only $q$ queries in any $\epsilon$ neighborhood in the space of classifier inputs.
	
	In this work we consider the most restrictive cases of $q=1$ and $q=2$. We find that different optimization algorithms are most effective depending on which of these restrictions we adhere to, and thus present two attacks. At a high level, both of these attacks follow a similar procedure to existing (white-box) universal attacks: we sample one or more inputs from a distribution, and give these inputs to an optimization algorithm that provides an update to the universal perturbation $\boldsymbol{\delta}$. However, in contrast to white-box approaches, our optimization algorithms do not rely on gradient information at these points, and to satisfy the undetectability restriction, we sample from our dataset without replacement and query each image at most $q$ times.
	
	\subsection{Loss Functions for Targeted and Untargeted Attacks}
	Building on the proposed black-box optimization techniques for query limited attacks, we now specify the choice of attacker's loss. For the case of untargeted attacks, we take the loss function in \eqref{eq:cross-entropy} to be the objective that the attacker seeks to maximize. 
	%
	For targeted attacks, the loss function is modeled as the \emph{difference} in losses between the target label logit $y_t \in \{1,\ldots,K\}$ and true label $y$ (this is also applicable in the case where the true label is not known, by simply choosing $y$ to be the next-largest logit besides the target class $y = \argmax_{k \neq y_t} C(x)_k$.) In this case, the loss function that the attacker optimizes is actually given by the \emph{difference} of cross-entropy terms
	{\small \begin{align}
		\label{eq:t_attack}
		&L(\mathbf{x}+\boldsymbol{\delta}, y)  =  
		\left ( \log\sum_{k=1}^K \exp(C(\mathbf{x})_k) - C(\mathbf{x})_{y} \right )\nonumber\\&- \left ( \log\sum_{k=1}^K \exp(C(\mathbf{x})_k) - C(\mathbf{x})_{y_t}  \right )\nonumber \\ &= C(\mathbf{x})_{y_t} -  C(\mathbf{x})_{y}
		\end{align}}
	The loss function minimizes cross-entropy loss for the target class and maximizes that of the highest prediction class other than the target.
	
	Despite its similarity to untargeted attacks, crafting targeted perturbations in black-box settings, particularly in attacks on networks with many output classes (large $K$), is a considerably more challenging task, as the perturbation must bring $\mathbf{x}$ into a region that is classified to a particular output class $y_t$, rather than any output class besides the true label $y$.
	

	\subsection{Dimensionality Reduction}
	The input spaces of many deep learning classifiers, and in particular deep image classifiers, are extremely high dimensional, which causes zeroth-order optimization approaches to be extremely query hungry and inefficient when applied to the input space directly. A frequently used tactic to alleviate this problem is to instead solve for the perturbation in a low-dimensional space.
	
	To this end, we employ a simple dimension-reduction technique, where instead of modeling the perturbation in the original input dimension $d\times d\times 3$, we do so in a smaller space of dimension $l \times l \times 3$, where $l \ll d$. This smaller perturbation is then repeated (tiled) across the image in a non-overlapping fashion until it is of size $d \times d \times 3$ (cutting off the final repetition of the rightmost and lowest tiles if $l$ does not divide $d$). We note that this technique is different from a common dimension reduction technique, also sometimes referred to as tiling, that first downsamples the image to a lower dimensional space, generates a perturbation for the downsampled image, and finally upsamples the perturbation into the original image space.
	
	This tiling technique leads to a trade-off between the low-dimensional perturbation size $l$ and the attack success rate. While a small $l$ makes black-box optimization easier and faster, it enforces a rigid structure on the perturbation due to tiling in the image space, thereby leading to a potentially lower attack success rate. On the other hand, as large $l$ allows for optimization in a larger space of perturbations, it potentially leads to higher attack success rates, at the cost of more difficult and query-inefficient optimization.

	\subsection{YOQO: Covariance Matrix Adaptation (CMA-ES)}
	\begin{algorithm*}[t]
		\caption{You Only Query Once with CMA-ES}
		\label{alg:yoqo}
		\begin{algorithmic}[1]
			\Procedure{\textsc{YOQO}}{$\{\mathbf{x}_i\}_{i=1}^{N}, \{y_i\}_{i=1}^{N}, L(\mathbf{x},y), \lambda, \sigma_0, \epsilon$}
			\noindent \State population size $\lambda$, initial step size $\sigma_0$, batch size $B$, maximum iterations $T$, $\ell_{\infty}$ bound $\epsilon$
			\noindent \State Initialize CMA-ES with step size $\sigma_0$, mean $\mu:=0$, covariance $\Sigma:=I$, population size $\lambda$, box constraints $\pm \epsilon$
			\noindent\For{$t = 1, \ldots, T$}
			\noindent\For{$j = 1, \ldots, \lambda$}
			\State $\mathbf{z}_j \sim \mathcal{N}(\boldsymbol{\mu}, \sigma^2 \bm{\Sigma})$
			\noindent\Comment{Draw perturbation tile from multivariate Gaussian}
			\noindent\State $\mathbf{\hat z}_j = $ Repeat tile $\mathbf{z}_j$ to match input image size
			\noindent\State Sample without replacement $(\mathbf{x}_{j,1},y_{j,1}),\ldots, (\mathbf{x}_{j,B}, y_{j,B})$
			\noindent\State $f_j = \frac{1}{B}\sum_{b=1}^B L(\mathcal{P}_{[0,1]}(\mathbf{x}_{j,b} + \mathbf{\hat z}_j), y_{j,b})$  \Comment{Compute loss function on batch of inputs}
			
			\EndFor
			\State $(\boldsymbol{\mu}, \sigma, \bm{\Sigma}) \leftarrow \text{CMAESUpdate}((\mathbf{z}_1, f_1), \ldots, (\mathbf{z}_\lambda, f_\lambda))$ \Comment{Use each tile's evaluation to update CMA-ES parameters}
			\State $j'=\argmax_{j\in\{1,\ldots, \lambda\}} f_j$ \Comment{Choose index of tile that maximizes loss}
			\State $\boldsymbol{\delta} = z_{j'}$ \Comment{Select tile that maximizes loss as current best perturbation}
			\EndFor
			\State \Return $\boldsymbol{\delta}$
			\EndProcedure
		\end{algorithmic}
	\end{algorithm*}
	
	The covariance matrix adaptation evolution strategy~(CMA-ES) \cite{hansen2016cma} is a popular and versatile derivative-free optimization algorithm that sees wide use in a range of fields and applications. Briefly, CMA-ES keeps track of a parameterized multivariate Gaussian with mean $\boldsymbol{\mu}$, step size $\sigma$ and covariance $\bm{\Sigma}$. At each iteration, a population of samples $\{\mathbf{z}_j\}_{j=1}^\lambda$ is drawn from $\mathcal{N}(\boldsymbol{\mu}, \sigma^2 \bm\Sigma)$. Each sample is evaluated (with no assumptions that this evaluation be differentiable or smooth), samples are ranked according to their evaluation, and the CMA parameters $\{\boldsymbol{\mu}, \sigma, \bm{\Sigma}\}$ are updated based on this ranking. The inner workings of an iteration of CMA-ES is quite complex; the full update is provided in the supplement.
	
	To apply CMA-ES to the problem of solving \eqref{eq:opt2}, we let each population sample $\mathbf{z}_j$ represent a perturbation tile. We evaluate $\mathbf{z}_j$ by tiling it to create a perturbation $\mathbf{\hat z}_j$ of the same size as a network input, perturbing each image in a minibatch of $B$ network inputs by adding $\mathbf{\hat z}_j$, and computing the loss function. Using these evaluations, the CMA-ES parameters are updated and the iteration's best perturbation (i.e. the one corresponding to the highest loss in the minibatch) is chosen. This iteration continues until we run out of images or any other desired stopping criteria is reached. The full procedure is listed in full in Algorithm~\ref{alg:yoqo}.
	
	CMA-ES has several favorable properties: it deals well with high variance by considering only the relative \emph{ranking} of samples within a population, rather than the raw objective values of those samples; it incorporates momentum-like mechanism via its evolution path; and it modifies the covariance $\bm{\Sigma}$ over time to avoid wasting queries in directions that do not affect samples' objective values. All of the hyperparameters of CMA-ES have prescribed values based on the dimension and box constraints of the problem, and practitioners are generally discouraged from deviating from this configuration. Nevertheless, in ~\cite{hansen2016cma} the authors suggest that, in settings with particularly high variance in objective values such as ours, a larger population size $\lambda$ may improve overall performance. Increasing this value allows us to better approach the high variance in \eqref{eq:opt2}, at the cost of more queries per CMA-ES iteration and slower convergence.
	
	\subsection{YOQT: Finite Difference Based Attacks}

	\begin{algorithm*}[t]
		\caption{You Only Query Twice: Finite difference based attacks}
		\label{alg:yoqt}
		\begin{algorithmic}[1]
			\Procedure{\textsc{YOQT}}{$\{\mathbf{x}_i\}_{i=1}^{N}$, $\{y_i\}_{i=1}^{N}$, $\bm{\delta}$, $L(\mathbf{x},y)$, $B, T, \mu, \epsilon, \eta, \{\mathbf{z}_j\}_{j=1}^{D}, D$}
			\State $\bm{\delta}$ is the perturbation, $B$ is the batch size, $T$ is the number of iterations, $\mu$ is the scaling for the perturbation, $\epsilon$ is the $\ell_{\infty}$ bound, $\eta$ is the step size, $\{\mathbf{z}_j\}$ are the vectors to be used for querying
			\State $\mathbf{g}_{avg}=0$
			\While{$t \leq T$}
			\State $\widehat{\mathbf{g}}=0$
			\For{$j=1, \cdots, J$}
			\State Sample without replacement $(\mathbf{x}_{j,1},y_{j,1}),\cdots, (\mathbf{x}_{j,B}, y_{j,B})$
			\Comment{Sampling B images}
			\State $\mathbf{g}=0$
			\Comment{Initializing the estimated gradient}
			\For{$b=1, \cdots , B$}
			\State Repeat tile $\mathbf{z}_j$ to match input image size
			\State {\small $L_1 = L\left(\mathcal{P}_{[0,1]}(\mathbf{x}_{j,b}+\mathcal{P}_{B(0,\epsilon)}^\infty(\bm{\delta}+\mu\mathbf{z}_{j}))\right)$}
			\Comment{Querying the model}
			\State {\small $L_2 = L\left(\mathcal{P}_{[0,1]}(\mathbf{x}_{j,b}+\mathcal{P}_{B(0,\epsilon)}^\infty(\bm{\delta} - \mu\mathbf{z}_{j}))\right)$ }
			\Comment{Querying the model}
			\State $\mathbf{g} = (\mathbf{g}*b+\frac{L_1 - L_2}{2\mu}\mathbf{z}_j)/(b+1)$
			\EndFor
			\State $\widehat{\mathbf{g}} = (\widehat{\mathbf{g}}*j+\mathbf{g})/(j+1)$
			\EndFor
			\State $\mathbf{g}_{avg}=0.5*\mathbf{g}_{avg}+0.5*\widehat{\mathbf{g}}$ 
			\State $\bm{\delta} = \mathcal{P}_{B(0,\epsilon)}^\infty\Big(\bm{\delta}+\eta*\textit{sign}(\mathbf{g}_{avg})\Big)$
			\Comment{Ensuring the perturbation is feasible}
			\EndWhile
			\State \Return $\bm{\delta}$
			\EndProcedure
		\end{algorithmic}
	\end{algorithm*}
	
	The second class of black-box optimization methods employed in this work are finite difference-based methods, which are built on explicit gradient estimation. We use this class of methods in a manner akin to stochastic projected gradient descent as depicted in \eqref{eq:pgd_update}; the main challenge being to estimate gradients with effective reduction of bias and variance. Not withstanding the bias and variance of finite difference based gradient estimation schemes, further constraining the query access to just $2$ black-box queries per image makes the universal adversarial perturbation problem at hand particularly challenging which is what our black-box method seeks to alleviate.
	
	With only black-box query access to the classifier, and while conforming to two queries per image limitation, the proposed method estimates  $\mathbf{\Bar{g}}^{(t)}$ as
	\begin{align}
	\label{eq:fd_gen}
	\widehat{\mathbf{g}}^{(t)}  &: = \dfrac{1}{BJ}  \sum_{j=1}^{J}\sum_{b=1}^{B} \frac{L(\mathbf{x}_{j,b}+\bm{\delta}^{(t)}+\mu \mathbf{z}_j, y_{j,b})}{2\mu}\mathbf{z}_j\nonumber\\&-\frac{L(\mathbf{x}_{j,b}+\bm{\delta}^{(t)}-\mu \mathbf{z}_j, y_{j,b})}{2\mu}\mathbf{z}_j,
	\end{align}
	where $\mathbf{z}_b$ denotes the direction along which the image is \emph{perturbed}, and $\mu$ is the smoothing parameter. In \eqref{eq:fd_gen}, at iteration $j$, the same directional perturbation $\mathbf{z}_j$ is used for the entire batch of samples $\{\mathbf{x}_{j,b}, y_{j,b}\}_{b=1}^{B}$, where $B$ denotes the batch size. Note that, in \eqref{eq:fd_gen} each image is only queried twice, and is discarded afterwards, thus never queried again.

	However, owning to just one direction of perturbation being used, this gradient estimate tends to be erroneous with high variance. In order to alleviate the issue, we repeat this procedure $J$ times, i.e., sampling a new batch $\{\mathbf{x}_{j,b}\}_{b=1}^{B}$ without replacement at each iteration $j$ and then observing the loss along the perturbation directions $\{ \pm\mathbf{z}_j\}$ per image, giving rise to the gradient estimate described in \eqref{eq:fd_gen}.
	Typical choices for directions $\{\mathbf{z}_j\}$ are canonical basis vectors, normal random vectors, and Fast Fourier Transform basis~(FFT) vectors.

	The gradient estimates given by finite difference methods are biased by default, controlled by the smoothing parameter \mbox{$\mu > 0$} in \eqref{eq:fd_gen}, where high values of  $\mu$ lead to a higher bias, while a lower $\mu$ can lead to an unstable estimator. 
	For the untargeted setting, we use low frequency basis vectors from the FFT basis, while for the targeted setting we use the canonical basis vectors as directions for perturbations.
	Note that, the universal perturbation $\bm{\delta}^{(t)}$ and the perturbation direction $\mathbf{z}_j$ are in the reduced dimension space of $l\times l\times 3$. Thus, before using them in the update in \eqref{eq:fd_gen} they are repeated (tiled) across the image so as to conform with the image dimension. 
	The above constitutes one iteration of the black-box finite difference based method and we run $T$ iterations of the method, where in every update of a projected gradient ascent step,  a momentum step is also incorporated as 
	\begin{align*}
	\mathbf{g}_{avg}^{(t)} = (\mathbf{g}_{avg}^{(t-1)} + \widehat{\mathbf{g}}^{(t)})/2.
	\end{align*}
	Finally, $\mathbf{g}_{avg}^{(t)}$ is used as a surrogate for $\mathbf{\Bar{g}}^{(t)}$ in \eqref{eq:pgd_update}. The proposed finite-difference based algorithm is outlined in Alg. \ref{alg:yoqt}.

	\section{Experiments}
	
	In this work, experimental results are drawn from attacks performed on classifiers trained on the commonly used datasets of MNIST \cite{mnist}, CIFAR-10\cite{cifar}, and ImageNet\cite{imagenet}. In all cases, we assume that input image pixel values lie in the range $[0,1]$, and subject $\delta$ to an $\ell_\infty$-norm bound of $\epsilon$, where $\epsilon = 0.3$ for attacks on MNIST classifiers, $\epsilon = 16/255$ for attacks on CIFAR-10 classifiers, and $\epsilon = 0.05 = 12.75/255$ for attacks on ImageNet classifiers. The MNIST and CIFAR-10 attacks target pretrained models that are frequently used to evaluate and benchmark adversarial attacks, whereas ImageNet attacks are performed against a variety of pretrained classifiers.
	
	The dataset used by the attack to query the classifier varies by experiment and is specified in each section. However, all of the following experiments use a fixed holdout set of images to evaluate the universal perturbation that is not queried during the attack. For MNIST and CIFAR-10 experiments, this holdout set is each dataset's test set, whereas for ImageNet experiments we fix a randomly chosen 10,000 image subset of the ImageNet validation set. Hyperparameters for each experiment were chosen by grid search and are provided in the supplement.
	
	
	
	All of our attacks were implemented and performed using the PyTorch deep learning framework \cite{paszke2019pytorch}. For the CMA-ES portion of the YOQO attack, we used pycma \cite{hansen2019pycma}, a Python implementation of CMA-ES continuously updated by the original creators of the CMA-ES algorithm that internally handles boundary constraints (i.e., ensures that $\ell_\infty$ constraints are satisfied by population sample draws in line 6 of Algorithm \ref{alg:yoqo}).
	
	\subsection{Untargeted Attacks}
	
	In the untargeted attack setting, we seek to generate a universal perturbation $\delta$ that will cause misclassification. In all results and tables referenced in this subsection, attacks are evaluated only on correctly classified images in the holdout set, and each reported attack success rate is a median over 5 performed attacks.
	
	\begin{table}[t]
		\centering
		\begin{tabular}{lllcc}
			\toprule
			&  &  & \multicolumn{2}{c}{Success Rate}   \\
			&  & Images & YOQO   & YOQT  \\ 
			Dataset & Classifier & Queried & ($q=1$)   & ($q=2$)  \\ 
			\midrule
			& & 500   & 15.1\% & 88.7\% \\
			MNIST & ConvNet & 1000  & 49.9\% & 89.2\% \\
			& & 60000 & 83.6\% & 90.3\% \\
			\midrule
			CIFAR-10 & ConvNet     & 50000 & 74.1\% & 75.9\% \\
			\midrule
			& \small ResNet-50    & 40000 & 67.45\% & 71.45\%  \\
			& \small ResNet-101   & 40000 & 60.95\% & 64.62\%  \\
			ImageNet & \small VGG-16       & 40000 & 67.45\% & 76.87\%  \\
			& \small DenseNet-161 & 40000 & 72.39\% & 64.08\%  \\
			& \small Inception v3 & 40000 & 21.95\% & 45.77\%  \\
			\bottomrule
		\end{tabular}
		\caption{Success rates for untargeted $\ell_\infty$ universal attacks on MNIST, CIFAR-10, and Imagenet classifiers}
		\label{tab:untargeted}
	\end{table}
	
	When attacking MNIST and CIFAR-10 classifiers in the untargeted setting, we use each dataset's training set to query the classifier, and attack with both YOQO and YOQT algorithms using a tile size of $7 \times 7 \times 1$ (dimension $49$) for MNIST and $4 \times 4 \times 3$ (dimension $48$) for CIFAR-10. Table~\ref{tab:untargeted} shows attack success rates from both of our attack algorithms on both datasets, exhibiting that, in all four experiments, the obtained universal perturbations successfully misclassify the large majority of inputs. In particular, attacks on the MNIST classifier were evaluated at several points during the attack, showing that the YOQT algorithm in particular is capable of finding a powerful universal adversarial perturbation after querying relatively few images.
	
	
	
	When attacking ImageNet classifiers, our algorithms used the remaining $40,000$ images in the ImageNet validation set that are not in our holdout set to query the network. In these experiments, we found that the most effective tile size depended on the algorithm used: $3 \times 3 \times 3$ (dimension 27) when attacking with the YOQO algorithm ($q=1$), and $8 \times 8 \times 3$ (dimension 192) when using YOQT ($q=2$). A variety of commonly used ImageNet classifiers were considered: including Inception v3, ResNet50, ResNet-101, VGG-16 with batch normalization, and DenseNet-161 \cite{inception, resnet, vgg, densenet}. Results of these experiments are shown in Table~\ref{tab:untargeted}. As expected, the YOQT attack is able to find more effective universal perturbations than the more restrictive YOQO attack, but against nearly all architectures, both algorithms are able to produce universal adversarial perturbations that misclassify more than 60\% of all inputs. The sole exception is the Inception v3 architecture, one of the most difficult ImageNet architectures to attack and often not considered even by white-box universal adversarial attacks \cite{chaubey2020universal}. Even in this case, the YOQT algorithm is capable of producing a perturbation that misclassifies nearly half of all inputs.

	\begin{table}[t]
		\centering
		\begin{tabular}{lcc}
			\toprule
			Attack & Dataset & Mean Success Rate  \\
			\midrule
			YOQT ($q=2$) & MNIST &  66.8\%  \\
			YOQO ($q=1$) & CIFAR-10 & 28.3\% \\
			YOQT ($q=2$) & CIFAR-10 & 51.9\%  \\
			\bottomrule
		\end{tabular}
		\caption{Success rates for $\ell_\infty$ universal targeted attacks on MNIST and CIFAR classifiers as evaluated on the holdout set.}
		\label{tab:mnist_cifar_targeted}
	\end{table}

	\begin{figure*}[t]
		\label{fig:untargeted}
		\centering
		\includegraphics[scale=0.3]{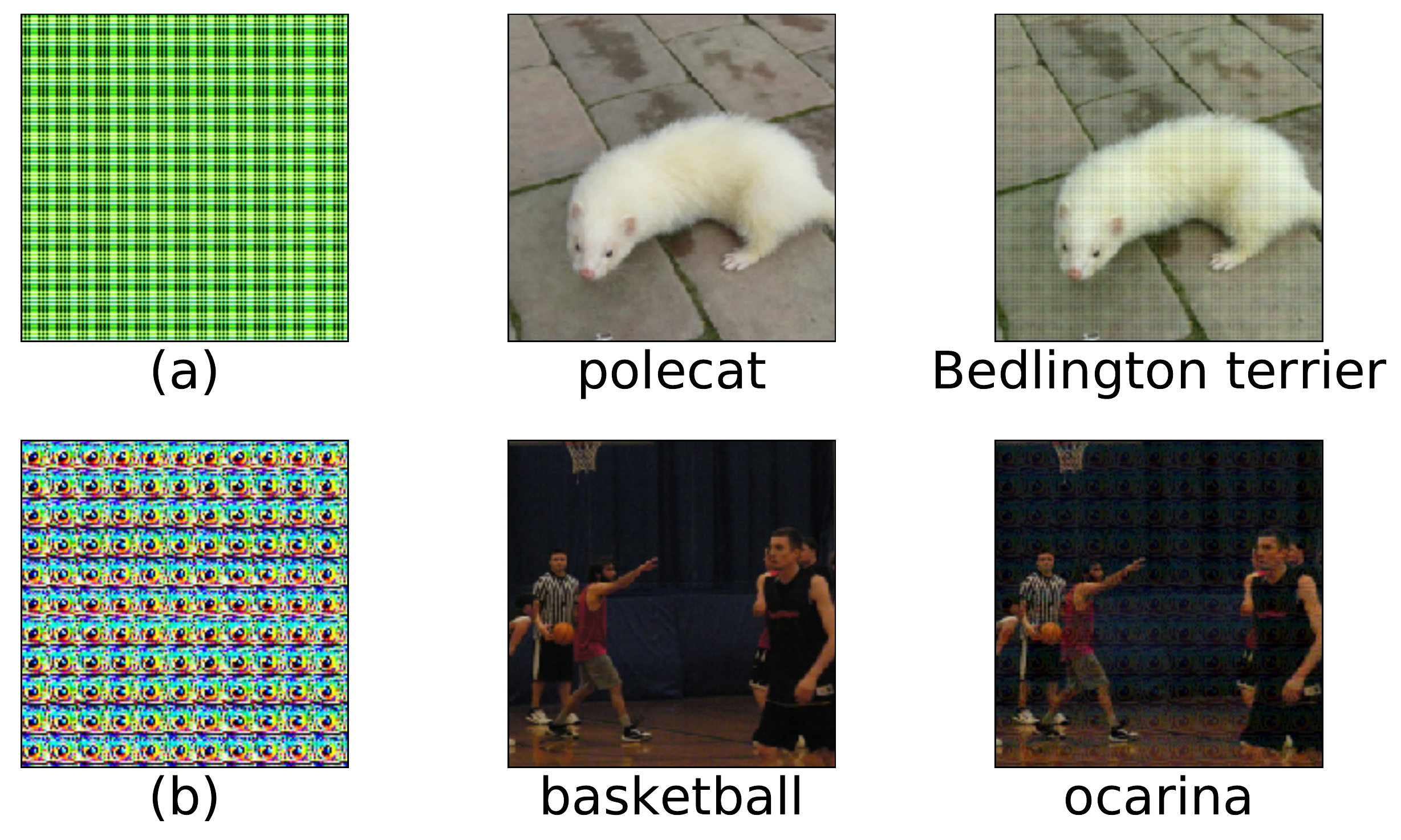} \hspace{0.5 in} \includegraphics[scale=0.3]{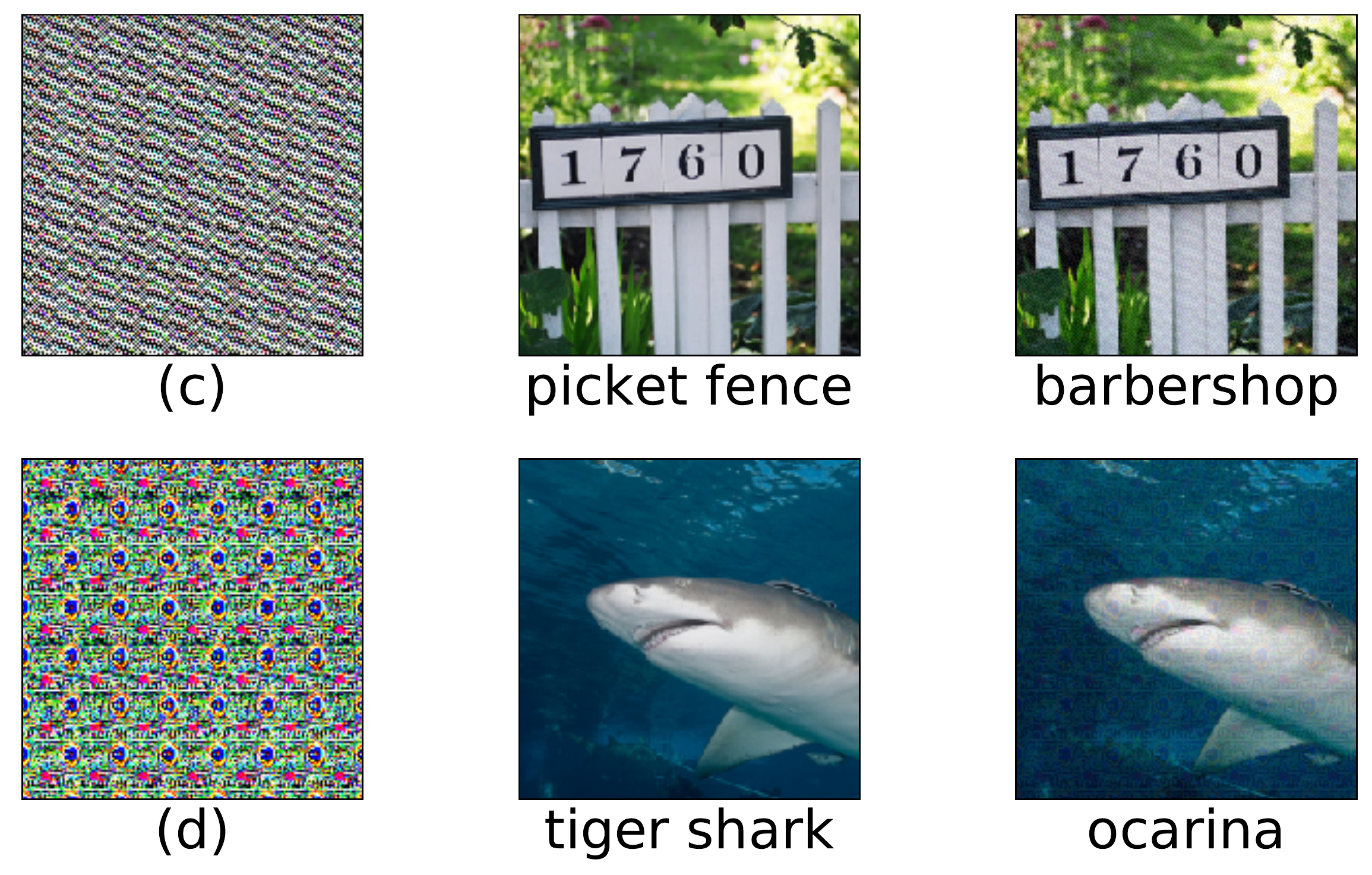}
		\caption{Four examples of universal black-box perturbations generated from our attacks on the ResNet-50 ImageNet classifier, with correctly classified clean images and incorrectly classified perturbed images. Perturbations are generated from (a) untargeted YOQO attack, (b) targeted YOQO attack, (c) untargeted YOQT attack, and (d) targeted YOQT attack.}
		
	\end{figure*}
	
	\subsection{Targeted Attacks}

	\begin{table*}[t]
		\centering
		\begin{tabular}{@{}cccccccccc@{}}
			\toprule
			\multirow{2}{*}{\textbf{Attack}} & \multirow{2}{*}{\textbf{Tile Size}} &
			\multicolumn{2}{c}{\textbf{Label $36$}  (Terrapin)} & \phantom{x} &
			\multicolumn{2}{c}{\textbf{Label $684$} (Ocarina)} & \phantom{x} &
			\multicolumn{2}{c}{\textbf{Label $699$} (Panpipe)} \\ 
			\cmidrule{3-4} \cmidrule{6-7} \cmidrule{9-10} && Queries & Success && Queries & Success && Queries & Success \\
			\midrule
			YOQO  & $20 \times 20 \times 3$ &- & -& & $4\times10^7$ & 20.57\%  & & - & -\\
			YOQT & $32 \times 32 \times 3$ & $4\times10^6$ & 51.92\%&& $4\times10^6$ & 44.45\% && $4\times10^6$ & 51.33\% \\
			\bottomrule
		\end{tabular}
		\caption{Success rates for targeted attacks on ResNet50 as evaluated on the holdout set}
		\label{tab:targeted}
	\end{table*}
	
	We also demonstrate that the attacks presented in this work are effective for devising targeted attacks. In this setting, we seek to influence the classifier to predict a particular target class $y_t$. Here, an attack is considered successful only if the unperturbed image is not classified as $y_t$ and the perturbed image is classified as $y_t$. Attack success rates in this section are evaluated on all images in the holdout set that are not originally classified as $y_t$, regardless of the ground truth label of the image.
	
	
	The problem of finding effective targeted attacks is significantly more difficult than the untargeted setting. This is reflected in several necessary changes to our algorithm; most notably, we find that the datasets used in the previous section to query the classifier are not always sufficiently large for our attacks to produce effective adversarial perturbations, and we thus draw from additional data sources when necessary. Since the targeted attack loss function \eqref{eq:t_attack} does not depend on the query image's true class label, any set of images, labeled or unlabeled, is appropriate for querying when crafting a targeted attack, even if these images do not fall into one of the classifier's expected classes. For targeted attacks on the MNIST classifier, we use additional data from the Extended MNIST (EMNIST) dataset \cite{emnist}, which includes just under $700,000$ images, while for ImageNet classifiers we use the ImageNet training set, which provides an additional 14 million images. For CIFAR-10 classifier attacks, we do not require additional data, and use the CIFAR-10 training set, as in the untargeted setting.
	
	Beyond larger training sets, targeted universal adversarial perturbations also required significantly different hyperparameter configurations. The full details of these configurations are left to the supplement, but we note here that targeted attacks generally failed when using small tile sizes, which may be too restrictive to contain universal targeted perturbations. Instead, we required much larger search spaces: for MNIST and CIFAR-10 targeted attacks, this meant not using any tiling dimension reduction techniques, and directly optimizing a perturbation of the same size as the original image, while for ImageNet classifier attacks, we found tile sizes of $20 \times 20 \times 3$ for YOQO and $32 \times 32 \times 3$ for YOQT to be most effective. Additionally, instead of relegating to only low-frequency FFT basis vectors as in the untargeted setting, here we find it useful to allow YOQT to use the entire FFT or the entire canonical basis.

	We experimented with targeted attacks on both MNIST and CIFAR-10 classifiers, testing each of the 10 classes in each dataset as the target class. Similar to the experimental protocol in our untargeted experiments, we performed 5 attacks on each of the 10 classes in each dataset, and collected the median success rate of each set of 5. The mean of these median success rates over all classes is reported Table~\ref{tab:mnist_cifar_targeted} (see supplementary material for per-class success rates), excepting for the YOQO targeted attack on MNIST, which we found was not able to successfully find targeted adversarial perturbations. In the remaining cases, results show that, despite the restrictive query limitation, YOQO is able to generate perturbations that classify more than 25\% of all images as a particular target class, whereas with two queries per image the YOQT attack is able to improve on this and successfully attack more than half of all images.
	
	
	In Table~\ref{tab:targeted}, we show the results of targeted attacks on ResNet-50, with attacks targeting several different classes for YOQT and one class for YOQO, due to the extraordinarily long runtime of the latter algorithm in this case. In all, our algorithms are able to find universal perturbations capable of forcing classification as one particular class for 20\% of all inputs when allowed to query each image once, or half of all inputs when allowed to query each image twice. Though the number of image queries required to achieve these attack success rates are admittedly outside of the realm of practicality for a real-world attack scenario, we emphasize that this is the first time that \textit{any} black-box methods have been shown to be capable of finding such perturbations.
	
	
	\section{Conclusion}
	This work presents, for the first time, two universal adversarial attacks that operate entirely in a black-box fashion, and demonstrated the ability of these algorithms to construct effective universal perturbations in a variety of experimental settings. Beyond this, these attacks adhere to severe restrictions on the allowed number of queries per image; through this, they are (unlike nearly all existing query-based black-box attacks) able to avoid clear query patterns that an alert defender may easily recognize as signs that they are under attack. Thus, in addition to being the first black-box attacks of their kind, the YOQO and YOQT attacks introduced here represent another step toward more realistic and realizable adversarial attacks on deep learning classifiers.

	\bibliography{yoqo}
	
	\newpage \phantom{blahblah} \newpage
	
	\onecolumn
	
	\appendix
	
	\section{Full CMA-ES Algorithm}
	
	\begin{algorithm*}[t]
		\caption{CMA-ES Update}
		\label{alg:fullcmaes}
		\begin{algorithmic}[1]
			\Procedure{\textsc{CMAESUpdate}}{step size $\sigma$, mean $\mu$, Dimension $n$, perturbation bound $\epsilon$, initial step size $\sigma_0$}
			\State Set hyperparameters $\lambda, s, w_{i=1,\ldots,\lambda}, c_m, c_\sigma, d_\sigma, c_c, c_1, c_s, s_{\text{eff}}$ as in \cite[Appendix A]{hansen2016cma}
			\State Initialize $\sigma := \sigma_0, \mu := \mathbf{0} \in \mathbb{R}^n, \Sigma := \mathbf{I} \in \mathbb{R}^{n \times n} $ 
			\For{$t = 1, \ldots, T$}
			\For{$j = 1, \ldots, \lambda$}
			\State $\mathbf{z}_j \sim \mathcal{N}(\boldsymbol{\mu}, \sigma^2 \bm{\Sigma})$ and enforce box constraints (see \cite{hansen2019pycma})
			\State Reshape and tile $\mathbf{z}_j$ from $\mathbb{R}^n$ to match image size $\mathbb{R}^{3 \times 224 \times 224}$
			\State Sample dataset without replacement $(\mathbf{x}_{j,1},y_{j,1}),\ldots, (\mathbf{x}_{j,B}, y_{j,B})$
			\State $f_j = \frac{1}{B}\sum_{b=1}^B L(\mathcal{P}_{[0,1]}(\mathbf{x}_{j,b} + \mathbf{z}_j), y_{j,b})$  \Comment{Compute loss function on batch}
			\EndFor
			\State Reindex sample-value pairs $(z_j, f_j)$ such that $f_1 \leq \ldots \leq f_j$ 
			\State $\langle y \rangle_w \leftarrow \sum_{j=1}^s w_j \frac{(z_j - \mathbf{\mu})}{\sigma}$
			\State $\mathbf{\mu} \leftarrow \mathbf{\mu} + \sigma \langle y \rangle_w$
			\State $\mathbf{p_{\sigma}} \leftarrow (1 - c_c) \mathbf{p_{\sigma}} + \sqrt{(2 - c_\sigma) c_\sigma s_{\text{eff}}} \mathbf{\Sigma}^{-1/2} \langle y \rangle_w$ 
			\State $\sigma \leftarrow \sigma \cdot \exp \left( \frac{c_\sigma}{d_\sigma} \left(\frac{\|\mathbf{p_{\sigma}}\|}{E[\|\mathcal{N}(0, I)\|]} - 1 \right) \right)$ 
			\State $p_c \leftarrow (1 - c_c) p_c + \sqrt{(2 - c_c) c_c s_{\text{eff}}} \langle y \rangle_w$
			\State $\mathbf{\Sigma} \leftarrow \left(1 - c_1 - c_\mu \sum_{j=1}^\lambda w_j\right)\mathbf{\Sigma} + c_1 p_c p_c^T + c_\mu \sum_{j=1}^\lambda w_j^\circ \mathbf{z}_j \mathbf{z}_j^T$ 
			
			\State $\boldsymbol{\delta} \leftarrow \mathbf{z}_{\hat j}, \text{ with } \hat j = \argmax_j f_j$
			\EndFor
			\State \Return $\boldsymbol{\delta}$
			\EndProcedure
		\end{algorithmic}
	\end{algorithm*}
	
	The full CMA-ES algorithm is given \cite[Appendix A]{hansen2016cma}; for completeness, the update to the CMA-ES parameters $\{\boldsymbol{\mu}, \sigma, \bm{\Sigma}\}$ (denoted CMAESUpdate in Algorithm~1) using notation consistent with this paper is given in Algorithm~\ref{alg:fullcmaes}. The full CMA-ES algorithm is also given in \cite[Appendix A]{hansen2016cma}; we note that the major discrepancies in notation are the mean and covariance of the distribution (we use $\mu$ and $\Sigma$, respectively, while \cite{hansen2016cma} uses $m$ and $C$), as well as the number of positive recombination weights (we use $s$ while \cite{hansen2016cma} uses $\mu$). Also note that the loss function is  we use both positive and negative recombination weights; originally called active CMA-ES or aCMA-ES~\cite{jastrebski2006improving}, this practice has become common enough that it is now considered a part of CMA-ES by default.
	
	\newpage
	
	\section{Experimental Details \& Hyperparameters}
	
	\begin{table*}[t]
		\centering
		\begin{tabular}{lcccc}
			\toprule
			& Tile & Population  & Batch  & Modified Initial  \\
			Experiment & Size & Size $\lambda$ & Size $B$ & Step Size $\sigma_0$  \\
			\midrule
			Untargeted MNIST      &   $7 \times 7 \times 1$  & 150 &  1 & -    \\
			Untargeted CIFAR-10   &   $4 \times 4 \times 3$  & 150 &  1 & -    \\ 
			Untargeted ImageNet   &   $3 \times 3 \times 3$  & 150 &  1 & -    \\ 
			Targeted CIFAR-10     &   $4 \times 4 \times 3$  & 100 & 10 & -    \\ 
			Targeted ImageNet     & $20 \times 20 \times 3$  & 500 & 20 & 0.06 \\ 
			\bottomrule
		\end{tabular}
		\caption{Hyperparameter configurations used in targeted and untargeted attacks with YOQO. In experiments where a modified initial step size is not specified, the prescribed CMA-ES default of $0.6\epsilon$ was used.}
		\label{yoqo_params}
	\end{table*}
	
	\begin{table*}[t]
		\centering
		\begin{tabular}{lcccccc}
			\toprule
			& Tile & Batch  & Smoothing  &  Step   \\
			Experiment & Size & Size $B$ & Parameter $\mu$ & Size $\eta$ & FFT Basis \\
			\midrule
			Untargeted MNIST      & $7 \times 7 \times 1$   & 10 & 0.0005 & 1     & Low-frequency \\
			Untargeted CIFAR-10   & $4 \times 4 \times 3$   & 10 & 0.0005 & 0.01  & Low-frequency \\ 
			Untargeted ImageNet   & $8 \times 8 \times 3$   & 10 & 0.0005 & 0.01  & Low-frequency \\ 
			Targeted MNIST        & $28 \times 28 \times 1$ & 10 & 0.0001 & 1     & Canonical \\
			Targeted CIFAR-10     & $32 \times 32 \times 3$ & 10 & 0.0001 & 0.01  & Canonical\\ 
			Targeted ImageNet     & $32 \times 32 \times 3$ & 10 & 0.0001 & 0.01  & Canonical \\ 
			\bottomrule
		\end{tabular}
		\caption{Hyperparameter configurations used in targeted and untargeted attacks with YOQT.}
		\label{yoqt_params}
	\end{table*}
	
	Hyperparameters used in each attack varied depending on the type of classifier attacked. Best configurations were determined via grid search and are detailed for each experiment in Tables~\ref{yoqo_params} (YOQO) and \ref{yoqt_params} (YOQT).
	
	\newpage
	
	\section{Full Targeted Attack Results}
	
	Per-class universal targeted attack results on MNIST and CIFAR-10 classifiers are given in Table~\ref{tab:mnist_appendix}.
	
	\begin{table*}[t]
		\centering
		\begin{tabular}{lc}
			\toprule
			MNIST & Success Rate \\
			Class & YOQT \\
			\midrule
			0       &  56.13\% \\
			1       &   1.92\% \\ 
			2       &  88.76\% \\ 
			3       &  90.14\% \\ 
			4       &  76.27\% \\ 
			5       &  81.23\% \\ 
			6       &  61.83\% \\ 
			7       &  79.75\% \\ 
			8       & 100.00\% \\ 
			9       &  32.11\% \\
			\bottomrule
		\end{tabular}
		\hspace{1.0 in}
		\begin{tabular}{lcc}
			\toprule
			CIFAR-10 & \multicolumn{2}{c}{Success Rate} \\
			Class & YOQO & YOQT \\
			\midrule
			0 airplane   &  2.57\% & 14.71\% \\
			1 automobile &  4.67\% & 16.03\% \\ 
			2 bird       & 71.17\% & 80.13\% \\ 
			3 cat        &  9.18\% & 55.22\% \\ 
			4 deer       & 45.07\% & 67.21\% \\ 
			5 dog        & 10.01\% & 52.25\% \\ 
			6 frog       & 45.59\% & 83.53\% \\ 
			7 horse      &  4.62\% & 13.99\% \\ 
			8 ship       & 38.76\% & 57.02\% \\ 
			9 truck      & 51.37\% & 78.92\% \\
			\bottomrule
		\end{tabular}
		
		\caption{Success rate by class for universal targeted attacks on MNIST (left table) and CIFAR-10 (right table) classifiers}
		\label{tab:mnist_appendix}
	\end{table*}

\end{document}